\documentclass{article}
\usepackage{spconf,amsmath,epsfig}
\usepackage{graphicx}
\usepackage{framed,multirow}
\usepackage{amssymb}
\usepackage{latexsym}
\usepackage{booktabs}
\usepackage{algorithm}
\usepackage{algorithmic}
\usepackage{pifont}

\let\OLDthebibliography\thebibliography
\renewcommand\thebibliography[1]{
  \OLDthebibliography{#1}
  \setlength{\parskip}{0pt}
  \setlength{\itemsep}{0pt plus 0.3ex}
}

\title{Uncovering the human motion pattern: \\ 
Pattern Memory-based Diffusion Model for Trajectory Prediction 
}% Following in the Footsteps: \\ Predicting Human Trajectories Using Cluster-based Memory}
\name{Yuxin Yang$^\dagger$, Pengfei Zhu$^\dagger$\thanks{$^\dagger$ Equal contribution.}, Mengshi Qi\textsuperscript{\ding{41}}\thanks{\textsuperscript{\ding{41}} Corresponding author.}, Huadong Ma}
\address{
Beijing Key Laboratory of Intelligent Telecommunications Software and Multimedia, \\
Beijing University of Posts and Telecommunications}

\begin{document}\sloppy
%\ninept
%
\maketitle
%
% 问题背景+强调当前主要问题，即?，导致?变得具有挑战性。
% 提出解决这一问题的方法，即一种新颖的?+detail。
% eg 方法描述中，明确指出主要特点；该表示包含?info，从而避免?；tradeoff acc & v; 强调通用性，即能够无缝推广到xxx。
%extensive exp

% 问题背景-指出问题-提出解决方案-明确方法特点-实验证明方法的有效性

\begin{abstract}
Human trajectory forecasting is a critical challenge in fields such as robotics and autonomous driving. Due to the inherent uncertainty of human actions and intentions in real-world scenarios, various unexpected occurrences may arise. To uncover latent motion patterns in human behavior, we introduce a novel memory-based method, named \textbf{M}otion \textbf{P}attern \textbf{P}riors \textbf{M}emory \textbf{Net}work. 
% 基于聚类的运动分布模式先验
Our method involves constructing a memory bank derived from clustered prior knowledge of motion patterns observed in the training set trajectories. 
We introduce an addressing mechanism to retrieve the matched pattern and the potential target distributions for each prediction from the memory bank, which enables the identification and retrieval of natural motion patterns exhibited by agents, subsequently using the target priors memory token to guide the diffusion model to generate predictions.
Extensive experiments validate the effectiveness of our approach, achieving state-of-the-art trajectory prediction accuracy. The code will be made publicly available.
% Furthermore, we design an addressing mechanism to compare the given trajectories with those in the memory bank, identifying and retrieving patterns of the agents and extracting their potential endpoint distributions. Ultimately, we employ this distribution to guide the diffusion model to generate future trajectories. 
\end{abstract}

\begin{figure*}[!t]
\centering
\includegraphics[width=1\linewidth]{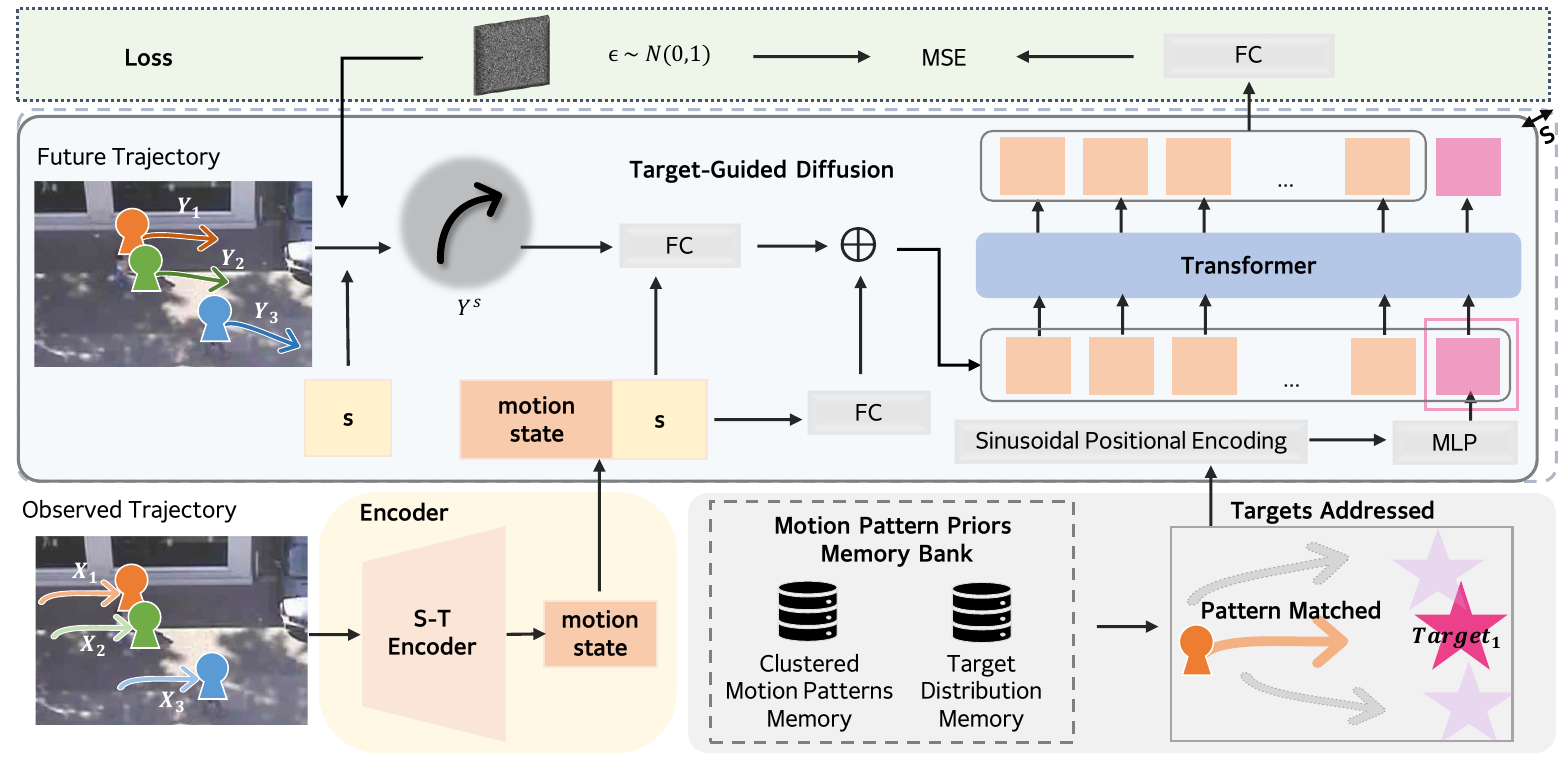}
\vspace{-5mm}
\caption{The overview of our proposed MP$^2$MNet method. It contains an encoder, the motion pattern priors memory bank, and a Transformer-based decoder. The encoder captures information to obtain the motion state representation. $S$ denotes the total diffusion step and $s$ denotes the $s^{th}$ step. $Y^s$ is corrupted $s$ steps by adding noise variable to ground-truth $Y^0$.
The decoder processes $Y^s$ along with motion state embedding, target priors memory token, and time embedding to generate the output. The training objective is to minimize the mean square error (MSE) loss between the output and the noise variable in the Gaussian distribution. This is achieved through target-guided diffusion generation for each iteration $s$ to optimize the network.
% for current iteration $s$ to optimize the network through the target-guided diffusion generation.
}
\vspace{-2mm}
\label{fig:overview}
\end{figure*}

\section{Introduction}
\label{sec:intro}
Human trajectory prediction plays a pivotal role in ensuring the safety and efficiency of applications such as autonomous driving systems~\cite{levinson2011towards} and social robots. 
% In traffic scenarios, humans are among the most encountered dynamic objects by self-driving cars. Accurately predicting the future motion paths of humans is essential for the planning and control of autonomous vehicles in such situations. 
However, predicting human trajectories is inherently challenging due to the unpredictable and subjective nature of human behavior.

% recent studies
To solve these challenges, significant strides have been achieved in research across different aspects, such as social interactions modeling~\cite{alahi2016social, huang2019stgat}. 
% In real-world applications, the effectiveness of a prediction system hinges on its adaptability to diverse scenarios, emphasizing its capability to accurately forecast a wide range of potential future trajectories in various dynamic situations.
To emphasize the capability of accurately forecasting a wide range of potential future trajectories in various situations,
many generative models are utilized for stochastic trajectory prediction to model the distribution of future trajectories, including generative adversarial networks (GANs)-based\cite{gupta2018socialgan}, conditional variational autoencoder (CVAE)-based~\cite{salzmann2020trajectron++,mangalam2020not,xu2022groupnet}, and diffusion-based~\cite{gu2022stochastic} methods.
%memory
Recently, memory networks have been used in the task of trajectory prediction. Many approaches create memory repositories with different contents.
Some methods~\cite{mangalam2020not, meng2022forecasting} store all trajectories or clusters of trajectories in the memory bank, and the selection for prediction simply relies on calculating cosine similarity between observed and stored trajectories. Other methods~\cite{xu2022remember} use memory mechanisms by storing starting and ending points and training a scoring network to select appropriate intentions.

% problem
% While numerous trajectory prediction methods have made substantial contributions, only a few are reliably applicable in real-world scenarios due to unnatural trajectory generation and limited exploration on uncertainties. 
While notable advancements have been made, unnatural trajectory generation and limited exploration of human motion uncertainties remain challenging.
% motivation: generative unnatural + memory pattern prior 
Because existing generative model-based methods lack reasonable guidance from motion pattern priors, and current memory-based approaches lack a systematic strategy to harness inherent uncertainties in motion pattern behavior. For instance, unexpected occurrences in real-world scenarios, like abrupt changes in human motion trajectories, are not adequately addressed and taken into consideration.

% our methods
To solve these problems, we introduce a novel memory approach that combines the strengths of the aforementioned stochastic methods and memory approaches to fully leverage the valuable probabilistic information and motion pattern priors. We employ a clustering method based on the motion trend of trajectories to obtain the human motion pattern priors and store them in the memory bank, which facilitates the prediction of a novel agent's recurring motion pattern given its observed motion state~\cite{ouellette1998habit}. Then we retrieve the matched pattern and the potential target distribution to obtain the target priors memory token, combined with the motion state as a condition to guide the learning of the reverse diffusion process. 
We optimize the
model using the evidence lower bound maximization method during the training, and sample the trajectories by denoising from a noise distribution during the inference. %Our method significantly enhances prediction performance. %leading to improved reliability and generalizability for practical applications without imposing a significant computational burden during the inference process.
Extensive experiments validate the effectiveness of our approach, achieving state-of-the-art results on ETH/UCY and  Stanford Drone datasets.
% starting destination x 
% motion pattern, clustering based on whole traj

In summary, the key contributions of our work can be summarized as follows:
% \par\textbf{(1)}~
1) We introduce a motion pattern prior memory bank to refine prediction results. This marks the first work of utilizing a clustering method to store human motion patterns with uncertainties and target distribution priors for prediction guidance.
2) We introduce a novel target-guided diffusion model in Motion Pattern Priors Memory Network (MP$^2$MNet). Using the matched motion pattern and target distribution, we can obtain the target priors memory token as guidance for the Transformer-based decoder in the reverse diffusion process to guide the diffusion model to generate various reasonable prediction results.
3) We conduct extensive experiments on multiple benchmark datasets, showcasing the superior performance of our method compared to state-of-the-art methods from recent years.

\section{Related works}\label{section:RelateWorks}
%\subsection{Trajectory prediction.} 
\textbf{Trajectory prediction:}
Human trajectory prediction methods can be classified into single-modality and multi-modality approaches. In single-modality approaches, early trajectory prediction research primarily relied on deterministic models like Markov processes~\cite{kumar2016ask} and recurrent neural networks~\cite{morton2016analysis}. Nevertheless, modern approaches emphasize modeling complex social interactions among agents. For example, Social-LSTM~\cite{alahi2016social} introduced a social pooling layer to capture agent interactions, extended by Social-GAN~\cite{gupta2018socialgan}. Additionally, attention-based methods are essential to capture critical interactions in crowded environments~\cite{huang2019stgat}. Other works have integrated scene understanding to extract global information, such as SS-LSTM~\cite{xue2018ss}. Trajectron++~\cite{salzmann2020trajectron++} established the connections between scene information and agent motion using graph structures, and MANTRA~\cite{marchetti2020mantra} combined memory mechanisms with scene images. MID~\cite{gu2022stochastic} is noteworthy as it introduced diffusion models to predict trajectories by modeling the process of human motion variation from indeterminate to determinate.
% We propose a single-modality method for human trajectory prediction called Motion Pattern Priors Memory Network (MP$^2$MNet), which employs encoder-decoder architecture.

% \subsection{Memory Networks.} 
% In~\cite{weston2014memory}, answers are stored in a knowledge base and retrieved based on the corresponding questions. Sukhbaata et al.~\cite{sukhbaatar2015end} employs an end-to-end memory network to answer questions.
\textbf{Memory Networks:}
Memory networks are commonly used in question-answering tasks. In recent years, memory networks have gained prominence in trajectory prediction tasks, demonstrating noteworthy advancements. Several methods have established memory repositories with diverse contents. For example, Ma et al.~\cite{ma2021continual} applied memory networks to trajectory prediction tasks, utilizing generative memory for continuous trajectory prediction. Mangalam et al.~\cite{mangalam2020not} utilized a memory network for single-agent trajectory prediction. 
% They proposed a memory repository that can store all trajectories and used cosine similarity to calculate the similarity between observed trajectories and select trajectories from the memory repository. 
Similarly, SHENet~\cite{meng2022forecasting} employed a similar approach by clustering trajectories within the same scene, filtering trajectories in the memory repository based on similarity, and utilizing multimodal information for human trajectory prediction. MemoNet~\cite{xu2022remember} focuses on target points, utilizing memory mechanisms to store starting and target points and training a scoring network for target point prediction. 

Different from these works, we propose a novel memory approach combining the strength of the diffusion model and using motion pattern priors to guide the trajectory generation.

\section{Our Method}
\label{sec:method}

\subsection{Problem Definition}
\label{ssec:problemdefinition}
The goal of human trajectory prediction is to forecast the future trajectories of multiple agents based on their historical motion paths. Mathematically, given a scenario where the observed historical trajectories are denoted as $\mathcal{Z}$, and where multiple agents are present, each with their observed historical trajectory data represented as $X_i=\{ x_i^t \}_{t=-T_{obs}+1}^0 \in \mathbb{R}^{T_{obs}\times 2}$, here $ i \in [1,2,\dots,N]$ represents the identifier of the observed agents in the environment, and $x_i^t\in \mathbb{R}^2$ represents the position of the $i^{th}$ agent in the scene at time $t$. Our method aims to predict the trajectories of multiple agents in the future period, denoted as %$\hat{Y}_i=\{\hat{y}_i^t\}_{t=1}^{T_{pred}}\in\mathbb{R}^{T_{pred}\times2}, i\in[1,2,\dots,N]$. 
$Y_i=\{y_i^t\}_{t=1}^{T_{pred}}\in\mathbb{R}^{T_{pred}\times2}, i\in[1,2,\dots,N]$. 
The predicted future trajectories should closely resemble the ground truth
% $Y_i=\{y_i^t\}_{t=1}^{T_{pred}}\in\mathbb{R}^{T_{pred}\times2}, i\in[1,2,\dots,N]$ 
as much as possible. 
In subsection~\ref{ssec:diffusion}, we denote $X$, $Y$ without the agent index i for the observed and predicted trajectory.

\subsection{Overall Network Architecture}
\label{ssec:subhead}
We present our MP$^2$MNet based on the diffusion model, which is a novel target-guided framework to formulate the trajectory prediction task as a reverse process of diffusion. %We generate the trajectory by gradually reducing the uncertainty from all walkable areas to the determinate prediction with a parameterized Markov chain.
%Training
% Given the initial ambiguous region
%inference
% Once the reverse process is trained, we can generate the plausible trajectories by a noise Gaussian through the reverse process.
To be specific, as depicted in Figure~\ref{fig:overview}, our approach consists of three parts: 1) an encoder network, 2) a motion pattern priors memory bank, and 3) a Transformer-based decoder. 

The encoder captures information from historical trajectories, producing the observed motion state embedding. Here we % follow the same setting as MID~\cite{gu2022stochastic} to 
apply the encoder of Trajectron++~\cite{salzmann2020trajectron++}. %for its ability to capture dynamic interactions between agents precisely. Encoders in other state-of-the-art methods can be used as well.
Using our motion pattern priors memory bank, we can retrieve the matched pattern and target distributions to generate target priors memory tokens as guidance. It combines with other information such as motion state embedding to serve as a condition for the Transformer-based decoder, designed to model Gaussian transitions in a Markov chain, 
% The decoder takes into account ground truth trajectories, noise variables, time embeddings, and positional relations, 
enabling the network to optimize and generate predictions through denoising. 
% This architecture has demonstrated promising results in capturing temporal dependencies in trajectories and optimizing prediction accuracy using mean square error (MSE) loss.

\subsection{Motion Pattern Priors Memory}
\label{ssec:memorybank}
For each cluster, we utilize the \textbf{Motion Pattern Priors Memory} module to construct the memory bank $\mathcal{Z}_{bank}$, with clustered trajectories and uncertainty value attached. Note that, the \textbf{Motion Pattern Priors Memory} module is used to refine prediction results, which means our generative predictions do not solely rely on the memory bank.
% As depicted in Fig. 1, 
This module can be summarized as three procedures,
i.e., 1) constructing the motion pattern priors memory bank, 2) retrieving the matched motion pattern with target distributions, and 3) generating corresponding target priors memory token.

% memory bank
% During the construction of the memory bank, we initially record all historical trajectories and subsequently cluster them to obtain the motion probability distribution based on their start and end points. Trajectories within the same cluster, characterized by similar starting and ending points, are consolidated to obtain motion probability distributions for representative trajectories.

\textbf{Memory Bank Initialization.} To efficiently improve memory usage, we utilize clustering to group similar trajectories based on their motion trends, forming clustered motion pattern distributions. These distributions are then stored in the memory bank along with their uncertainty values. 
To be specific, for all trajectories in the training set, we represent their trajectories as a set $\mathcal{Z}$, 
%which includes the start point $x^{-T_{obs}+1}$, 
with targets $y^{T_{pred}}$ included. 
Then we use K-means clustering to group similar trajectories considering the whole motion trend in the $\mathcal{Z}$. 
%with similar start point $x^{-T_{obs}+1}$ and the target $y^{T_{pred}}$ 
Assume we can obtain $K$ clusters, representing a total of $K$ motion pattern distribution priors, formulated as $\{\mathcal{N}\}_0^{K}$, of which $\mu_i$ and ${\sigma_i}^2$ represent 2D trajectories and uncertainty value of agent $i$. Corresponding target distributions of each cluster can be formulated as $\{\rho^{T_{pred}}\}_0^{K}\}$. Finally, we can obtain the motion pattern priors memory bank $\mathcal{Z}_{bank}=\{\{\mathcal{N}\}_0^{K},
% \{x_{past}^{-T_{obs}+1}\}_0^{K},
\{\rho^{T_{pred}}\}_0^{K}\}$, which can be used then as priors memory.

\textbf{Trajectory Addressing.} Given the past trajectory $X_i$ of agent $i$, we first use
% the starting point $x^{-T_{obs}+1}$ and the endpoint $x^{0}$ of 
$X_i$ to obtain the matched motion pattern priors $\mathcal{N}_i$ with ($\mu_i$, $\sigma_i$) from the memory bank by selecting the one with minimal Gaussian negative log-likelihood score.
% To obtain the matched pattern $\mathcal{N}_i$ with $\mu_i$ and $\sigma_i$, we tried to minimize the Gaussian negative log-likelihood loss.
% $\mu$ are treated as samples from Gaussian distributions with expectations and variances retrieved from the memory bank. 

For each motion pattern $\mathcal{N}_j$, assuming the provided trajectory $X_i$ follows a Gaussian distribution with mean $\mu_j$ and variance ${\sigma_j}^2$, we can calculate the Gaussian Negative Log-likelihood (NLL) score as follows:
% We denote $N$ sampled full trajectories from $\mathcal{N}_i$ as $\mathcal{T}$, and the corresponding past portion of $\mathcal{T}$ as $\mathcal{H}$, $\mathcal{H} = \{h_1, h_2, \ldots, h_N\}$.
% For each sampled historical trajectory $h_i$, 

% gaussian loss var:方差 没const
% \begin{equation}
% \text{NLL}(h_i, X_i) = -\frac{1}{T}\sum_{t=1}^{T}\log p(h_i(t) | x_i^t)
% \end{equation}
\begin{equation}
S_{NLL}=\frac{1}{2}\left(\log (\max (\text { $\sigma_j$, $\epsilon$ }))+\frac{(\text { $X_i$ }- \text { $\mu_j$ })^2}{\max (\text { $\sigma_j$, $\epsilon$ })}\right),%+\text { const. }
\end{equation}
where $X_i$ represents the provided past trajectory, $\mu_j$ corresponds to the trajectory associated with motion pattern $\mathcal{N}_j$ stored in the memory bank, and $\sigma_j$ represents the uncertainty of $\mu_j$. The parameter $\epsilon$ is a hyper-parameter introduced for stability. %$T$ is the length of the past trajectory, $h_i(t)$ and $x_i^t$ are the corresponding positions at time $t$ in $h_i$ and $X_i$, $p(h_i(t) | x_i^t)$ represents the conditional probability of observing $h_i(t)$ given $x_i^t$.
% $f(\cdot)$ can be a MLP,

The pattern $\mathcal{N}_i$ with $\mu_i$ and $\sigma_i$, which yields the minimum Gaussian NLL score, is considered the optimal match for the current agent $i$:
\begin{equation}
\mathcal{N}_i = \arg\min_{\mu_i, \sigma_i \in \mathcal{N}} S_{NLL}(X_i, \mu_i, \sigma_i).
\end{equation}
Subsequently, we can retrieve the potential target distribution $\rho_i^{T_{pred}}$ associated with the matched pattern from the memory bank, which is employed to guide the diffusion process for generating future trajectories.

\begin{table*}[htbp]
  \centering
  \caption{Quantitative results on the Stanford Drone dataset with Best-of-20 strategy in ADE/FDE metric. $\downarrow$ represents that lower is better. The best results are highlighted in bold.}
  \vspace{+1mm}
    \begin{tabular}{l|ccccc|c}
    \toprule
    Time  & Social-LSTM~\cite{alahi2016social} & Social-GAN~\cite{gupta2018socialgan} & Trajectron++~\cite{salzmann2020trajectron++} & MANTRA~\cite{marchetti2020mantra} & GroupNet+CVAE~\cite{xu2022groupnet} & MP$^2$MNet \\
    \midrule
    4.8s  & 31.19/56.97 & 27.23/41.44 & 19.30/32.70 & 8.96/17.76 & 9.31/\textbf{16.11} & \textbf{8.89}/16.45 \\
    \bottomrule
    \end{tabular}%
  \label{sdd}%
\end{table*}%

\textbf{Target Priors Memory Token Generation.}
Using the obtained potential target distribution $\rho_i^{T_{pred}}$ of the agent $i$ through the trajectory addressing mechanism, we use sinusoidal position encoding and MLP to convert 2D target positions into target embedding, used as an additional target prior for prediction refinement:
%  Position Encoding sin/cos的原公式
% \begin{equation}
% t = f(\text{PE}(\rho_i^{T_{pred}})) 
% ,\end{equation}
% where $\rho_i^{T_{pred}}$ is the memory target position of agent $i$, $\text{PE}(\cdot)$ represents sinusoidal position encoding, and $f(\cdot)$ denotes MLP. 
\begin{equation}
\label{eq:tem-emb}
    \begin{aligned}
    % \mathcal{R}
        \mathbf{h}_{2j}   &= f(\sin\left(\rho_i^{T_{pred}} / \lambda^{2j/D}\right)), \\ 
        \mathbf{h}_{2j+1} &= f(\cos\left(\rho_i^{T_{pred}} / \lambda^{2j/D}\right)),
    \end{aligned}
\end{equation}
where $\rho_i^{T_{pred}}$ represents the target position of agent $i$, $\mathbf{h}\in\mathbb{R}^{D}$ represents the target priors memory token, $j$ is the dimension index, and $D$ is the dimension of the embedding $\rho_i^{T_{pred}}$. $\lambda$ denotes the max period of the sinusoidal function. The wavelengths form a geometric progression from $2\pi$ to $10000\cdot2\pi$. And $f(\cdot)$ denotes MLP.

\subsection{Target-guided Diffusion Model}
\label{ssec:diffusion}
As depicted in Figure~\ref{fig:overview}, we utilize the Transformer as the decoder following the previous work~\cite{gu2022stochastic}. We denote that $\mathbf{Y}^s$ is corrupted by adding a noise variable for $s$ times to the ground truth trajectory $\mathbf{Y}^0$. 

Then we take $\mathbf{Y}^s$, the target priors memory token $\mathbf{h}$ combined with the motion state embedding denoted as $\mathbf{G}$, and time embedding as the input of the Transformer-based decoder for the reverse diffusion process.

The decoder is trained to generate trajectories from Gaussian noise conditioned on the information including $\mathbf{G}$ at each step of the denoising process. The reverse diffusion process progressively diminishes uncertainties across all accessible regions, ultimately leading to specific predictions using a parameterized Markov chain.
% In this part, our model generates the trajectory by using target guidance retrieved from the motion pattern prior memory to gradually reduce the uncertainties from all walkable areas to the determinate prediction with a parameterized Markov chain.

\textbf{Target-guided Diffusion.}
The reverse diffusion process is the joint distribution $p_\theta \left(\mathbf{Y}^{\left(0:S\right)} \mid \mathbf{G} \right)$ conditioned on $\mathbf{G}$, defined as a Markov chain with learned Gaussian transitions that begins with $p\left(\mathbf{Y}^S\right)=\mathcal{N}\left(\mathbf{Y}^S;\mathbf{0, I}\right)$:
{\small
\begin{equation}
p_\theta\left(\mathbf{Y}^{\left(0: S\right)}\right):=p\left(\mathbf{Y}^S\right) \prod_{s=1}^S p_\theta\left(\mathbf{Y}^{s-1} \mid \mathbf{Y}^s, \mathbf{G}\right),
\end{equation}
\begin{equation}
p_\theta\left(\mathbf{Y}^{s-1} \mid \mathbf{Y}^s, \mathbf{G}\right):=\mathcal{N}\left(\mathbf{Y}^{s-1} ; \boldsymbol{\mu}_\theta\left(\mathbf{Y}^s, s, \mathbf{G}\right), \boldsymbol{\Sigma}_\theta\left(\mathbf{Y}^s, s\right)\right),
\end{equation}
}
where $s$ denotes the diffusion step, $\theta$ represents our target-guided diffusion model's parameter, and $\boldsymbol{\Sigma}_\theta\left(\mathbf{Y}^s, s\right)$ equals to $\beta_s \mathbf{I}$. $\beta_s$ is the variance at the denoising step $s$, controlling the extent of added noise.

The forward diffusion process is a Markov chain that gradually adds Gaussian noise to raw trajectory data for $S$ steps according to a uniformly increasing variance schedule $\beta_1$,...,$\beta_S$, which constrains the level of noise injection. 
We can formulate the approximate posterior %$q\left(\mathbf{Y}^{(1: S)} \mid \mathbf{Y}^0\right)$ 
as: 
%\eta --> s; 小s denoising step
% y_(1:H) --> {Y}^(1:S); 大S denoising step总数 （100）
% Y^0 : gt
{\small
\begin{equation}
q\left(\mathbf{Y}^{(1: S)} \mid \mathbf{Y}^0\right) := \prod_{s=1}^S q\left(\mathbf{Y}^{s} \mid \mathbf{Y}^{s-1}\right),
\end{equation}
\begin{equation}
q\left(\mathbf{Y}^{s} \mid \mathbf{Y}^{s-1}\right)=\mathcal{N}\left(\mathbf{Y}^{s} ; \sqrt{1-\beta_s} \mathbf{Y}^{s-1}, \beta_s \mathbf{I}\right).
\end{equation}
}
% where $\beta_s$ is the variance at the denoising step $s$. 
Using the notation $\alpha_s:= 1 - \beta_s$ and $\bar{\alpha}_s:=\prod_{m=1}^s \alpha_m$, we can obtain:
\begin{equation}
q\left(\mathbf{Y}^{s} \mid \mathbf{Y}^0\right)=\mathcal{N}\left(\mathbf{Y}^{s}; \sqrt{\bar{\alpha}_s} \mathbf{Y}^{0},\left(1-\bar{\alpha}_s\right) \mathbf{I}\right).
\end{equation}
When the total number of denoising step $S$ is large enough, $q(\mathbf{Y}^S))$ approximates to $\mathcal{N}(\mathbf{Y}^S;0,\mathbf{I})$, where $\mathcal{N}$ is a Gaussian distribution.

\textbf{Training Objective.}
Our target-guided diffusion model performs training by optimizing the variational lower bound.
% maximizing the log-likelihood of the predicted trajectories given the ground truth
%$E\left[\log p_{\theta}(\mathbf{Y}^0)\right]$. 
As the exact log-likelihood is intractable,
we use the evidence lower bound maximization method and minimize the KL divergence. We can use KL divergence to directly compare $p_\theta\left(\mathbf{Y}^{s-1} \mid \mathbf{Y}^s, \mathbf{G} \right)$ 
against forward process posteriors:%, which are tractable when conditioned on $\mathbf{Y}^s$ and $\mathbf{G}$:
%elbo推kl divergence公式
{\small
\begin{equation}
\begin{gathered}
\mathcal{L}=\mathbb{E}_{q, s}\left[D_{K L}\left(q\left(\mathbf{Y}^{s-1} \mid \mathbf{Y}^s, \mathbf{Y}^0\right) \| p_\theta\left(\mathbf{Y}^{s-1} \mid \mathbf{Y}^s, \mathbf{G}\right)\right)\right] \\
=\mathbb{E}_{q, s}\left[D_{K L}\left(\mathcal{N}\left(\mathbf{Y}^{s-1} ; \boldsymbol{\tilde{\mu}_s}, \boldsymbol{\Sigma}_q(s) \| \mathcal{N}\left(\mathbf{Y}^{s-1} ; \boldsymbol{\mu}_\theta, \boldsymbol{\Sigma}_{(s)}\right)\right)\right]\right. \\
=\mathbb{E}_{q, s}\left[\left\|\boldsymbol{\mu}_\theta\left(\mathbf{Y}^s, s, \mathbf{G}\right)-\boldsymbol{\tilde{\mu}}_s\right\|_2^2\right],
\end{gathered}
\end{equation}
}
where $\boldsymbol{\mu}_\theta$ and $\boldsymbol{\tilde{\mu}}_s$ are calculated using the reparameterization trick as:
{\small
\begin{equation}
\begin{aligned}
\tilde{\boldsymbol{\mu}}_s\left(\mathbf{Y}^s, \mathbf{Y}^0\right) & =\frac{\sqrt{\bar{\alpha}_{s-1}} \beta_s}{1-\bar{\alpha}_s} \mathbf{Y}^0+\frac{\sqrt{\alpha_s}\left(1-\bar{\alpha}_{s-1}\right)}{1-\bar{\alpha}_s} \mathbf{Y}^s, \\
\tilde{\beta}_s & =\frac{1-\bar{\alpha}_{s-1}}{1-\bar{\alpha}_s} \beta_s \mathbf{I},
\end{aligned}
\end{equation}
\begin{equation}
\boldsymbol{\mu}_\theta\left(\mathbf{Y}^s, s, \mathbf{G}\right)=\frac{1}{\sqrt{\alpha_s}}\left(\mathbf{Y}^s-\frac{\beta_s}{\sqrt{1-\bar{\alpha_s}}} \epsilon_\theta\left(\mathbf{Y}^s, s, \mathbf{G}\right)\right),
\end{equation}
}
where $\epsilon_\theta(\cdot)$ denotes the noise predictor of our target-guided diffusion model for trajectory generation.
We optimize the network through diffusion generation by performing mean square error (MSE) loss between the output and a noise variable in standard Gaussian distribution for the current iteration $s$, following the work~\cite{gu2022stochastic}:
% optimization problem becomes:
\begin{equation}
\mathcal{L}_{M S E}(\theta, \phi)=\mathbb{E}_{\boldsymbol{\mathbf{z}}, \mathbf{Y}^0, s}\left\|\boldsymbol{\mathbf{z}}-\epsilon_{(\theta, \phi)}\left(\mathbf{Y}^s, s, \mathbf{G}\right)\right\| 
,\end{equation}
where $\theta$ and $\phi$ are parameters of the target-guided diffusion model and encoder respectively, and $\mathbf{z} \sim \left(\mathbf{\left(0,I\right)}\right)$.

\textbf{Inference.}
% Inverse
During the reverse process, with the reparameterization, we use the DDPM sampling technique to repeatedly denoise $Y^S$ to $Y^0$ by using the equation below for S steps as:
% we generate the trajectories from $y_K$ to $y_0$ as:
\begin{equation}
\mathbf{Y}^{s-1}=\frac{1}{\sqrt{\alpha_s}}\left(\mathbf{Y}^s-\frac{\beta_s}{\sqrt{1-\bar{\alpha_s}}} \epsilon_\theta\left(\mathbf{Y}^s, s, \mathbf{G}\right)\right)+\sqrt{\beta_s} \mathbf{z}.
\end{equation}
% where $\mathbf{z} \sim \mathcal{N}(\mathbf{(0,I)})$.

% \subsection{Target-guided token Generation}
% % The above-mentioned \textbf{MP$^2$MNet} yields supervised end-to-end training. Subsequently, we employ this distribution to guide the inverse diffusion process, ultimately generating precise trajectories.
% As depicted in Figure~\ref{fig:overview}, we utilize the Transformer as the decoder following the previous work~\cite{gu2022stochastic}. 
% % During the training process, we introduce ground-truth target trajectories with Gaussian blurring into the diffusion process. During inference, we employ the negative log-likelihood (NLL) loss to select the most similar trajectory pattern within the memory bank and extract the endpoint distribution associated with that pattern. 
% Using the obtained potential target distribution $\rho_i^{T_{pred}}$ of the agent $i$ through the trajectory addressing mechanism, we use sinusoidal position encoding and MLP to convert 2D target positions into target embeddings, used as an additional reference. $y_k$ is corrupted $k$ times by a noise variable from the ground truth trajectory $y_0$. Then we take $y_k$, target embedding, state embedding, and time embedding as the input of the Transformer-based decoder for the reverse diffusion process:
% \begin{equation}
% \text{target} = f(\text{PE}(\rho_i^{T_{pred}})) 
% ,\end{equation}
% where $\rho_i^{T_{pred}}$ is the memory target position of agent $i$, $\text{PE}(\cdot)$ represents sinusoidal position encoding, and $f(\cdot)$ denotes MLP. 

\section{Experiments}
\label{sec:experiments}

% \begin{table*}[htbp]
%   \centering
%   \caption{}
%     \begin{tabular}{l|ccccc}
%     \toprule
%     Time  & Social-LSTM~\cite{alahi2016sociallstm} & Social-GAN~\cite{gupta2018socialgan} & Trajectron++~\cite{salzmann2020trajectron++} & GroupNet+CVAE~\cite{xu2022groupnet} & MP$^2$MNet \\
%     \midrule
%     4.8s  & 31.19/56.97 & 27.23/41.44 & 19.30/32.70 & 9.31/16.11 & 8.89/16.45 \\
%     \bottomrule
%     \end{tabular}%
%   \label{sdd}%
% \end{table*}%

{\small
\begin{table*}[htbp]
  \centering
  \caption{Comparison of state-of-the-art methods on ETH/UCY datasets. $\downarrow$ represents that lower is better. We report ADE and FDE for predicting future 12 frames in meters. $\dagger$ represents our reproduction results. ‘AVG’ means the average result over 5 subsets. The best-of-20 is adopted for evaluation. The best results are highlighted in bold while the second-best results are underlined. }
  \vspace{+1mm}
  \resizebox{\linewidth}{!}{
    \tiny 
    \begin{tabular}{c|ccccc|c}
    \toprule
    \multirow{2}[4]{*}{Methods} & \multicolumn{6}{c}{Evaluation metrics: ADE$\downarrow$ / FDE$\downarrow$ (in meters)} \\
\cmidrule{2-7}          & ETH   & HOTEL & UNIV  & ZARA1 & ZARA2 & AVG \\
    \midrule
    LSTM  & 1.01 / 1.94  & 0.60 / 1.34 & 0.71 / 1.52  & 0.41 / 0.89 & 0.31 / 0.68 & 0.61 / 1.27 \\
    S-LSTM~\cite{alahi2016social} & 0.75 / 1.38 & 0.61 / 1.40 & 0.58 / 1.03 & 0.42 / 0.70 & 0.43 / 0.71 & 0.56 / 1.05 \\
    % Social Attention\cite{8460504} & 1.39 / 2.39 & 2.51 / 2.91 & 1.25 / 2.54 & 1.01 / 2.17 & 0.88 / 1.75 & 1.41 / 2.35 \\
    SGAN~\cite{gupta2018socialgan}  & 0.87 / 1.62 & 0.67 / 1.37 & 0.76 / 1.52 & 0.35 / 0.68 & 0.42 / 0.84 & 0.61 / 1.21 \\
    STGAT~\cite{huang2019stgat} & 0.68 / 1.33 & 0.38 / 0.72 & 0.56 / 1.21 & 0.34 / 0.69 & 0.29 / 0.60 & 0.45 / 0.91 \\
    MANTRA~\cite{marchetti2020mantra} & 0.70 / 1.76 & 0.28 / 0.68 & 0.51 / 1.26 & 0.25 / 0.67 & 0.20 / 0.54 & 0.39 / 0.98 \\
    % pecnet traj++*
    %\midrule
    % CARPe\cite{huang2019stgat} & 0.68 / 1.33 & 0.38 / 0.72 & 0.56 / 1.21 & 0.34 / 0.69 & 0.29 / 0.60 & 0.45 / 0.91 \\
    TPNMS~\cite{liang2021temporal} & 0.52 / 0.89 & 0.22 / 0.39 & 0.55 / 1.13 & 0.35 / 0.70 & 0.27 / 0.56 & 0.38 / 0.73 \\
    %Temporal Pyramid Network for Pedestrian Trajectory Prediction with Multi-Supervision
    Social-Implicit~\cite{mohamed2022social} & 0.66 / 1.44 & 0.20 / 0.36 & 0.31 / 0.60 & 0.25 / 0.50 & 0.22 / 0.43 & 0.33 / 0.67 \\
    GroupNet~\cite{xu2022groupnet} & 0.46 / \underline{0.73} & \textbf{0.15} / \textbf{0.25} & 0.26 / 0.49 & \underline{0.21} / \textbf{0.39} & \underline{0.17} / \underline{0.33} & \underline{0.25} / \textbf{0.44} \\
    % MID~\cite{gu2022stochastic} & 0.54 / 0.82 & 0.20 / 0.31 & 0.30 / 0.57 & 0.27 / 0.46 & 0.20 / 0.37 & 0.30 / 0.51 \\
    MID$^\dagger$~\cite{gu2022stochastic} & \underline{0.42} / 0.75 & 0.18 / 0.32 & \underline{0.23} / \underline{0.46} & \textbf{0.20} / \underline{0.40} & \underline{0.17} / 0.34 & \underline{0.25} / 0.46 \\
    \midrule
    % MP$^2$MNet(w/o memory) & \underline{0.42} / 0.75 & 0.18 / 0.32 & \underline{0.23} / \underline{0.46} & \textbf{0.20} / \underline{0.40} & \underline{0.17} / 0.34 & \underline{0.25} / 0.46 \\
    % MP$^2$MNet(w/ memory) & \textbf{0.39 / 0.69} & \underline{0.16} / \underline{0.27} & \textbf{0.21} / \textbf{0.43} & 0.23 / 0.48 & \textbf{0.16} / \textbf{0.32} & \textbf{0.23} / \textbf{0.44} \\
    Ours (w/o memory) & 0.44 / 0.76 & 0.18 / 0.33 & \underline{0.23} / 0.47 & 0.25 / 0.52 & 0.19 / 0.41 &  0.26 / 0.50 \\
    % MP$^2$MNet(w/ memory) & \textbf{0.39 / 0.69} & \underline{0.16} / \underline{0.27} & \textbf{0.21} / \textbf{0.43} & 0.23 / 0.48 & \textbf{0.16} / \textbf{0.32} & \textbf{0.23} / \textbf{0.44} \\
    Ours (w/ memory) & \textbf{0.39 / 0.69} & \underline{0.16} / \underline{0.27} & \textbf{0.21} / \textbf{0.43} & 0.23 / 0.48 & \textbf{0.16} / \textbf{0.32} & \textbf{0.23} / \textbf{0.44} \\
    \bottomrule
    \end{tabular}
}%
  \label{ethucy}%
\end{table*}%
}
% \vspace{-3mm}

\subsection{Experimental Setup}

\noindent\textbf{Datasets.} We evaluate the effectiveness of our MP$^2$MNet on two commonly used benchmarks for human trajectory prediction tasks: ETH/UCY~\cite{pellegrini2010improving,lerner2007crowds}, and Stanford Drone Dataset(SDD)~\cite{robicquet2016learning}. ETH/UCY includes positions of pedestrians in the world coordinates from 5 scenes: ETH, HOTEL, UNIV, ZARA1, and ZARA2. We evaluate our model on this dataset using the same approach as in previous works~\cite{gupta2018socialgan, huang2019stgat}, employing the leave-one-out method. %: training on four datasets and testing on the remaining one.
SDD is captured in a university campus environment %and comprises various entities and scenes observed 
from a bird's-eye view.
For both datasets, we use the data from the past 8 frames (3.2 seconds) to predict the trajectory for the future 12 frames (4.8 seconds).

\noindent\textbf{Evaluation Metrics.} 
Evaluation metrics include ADE and FDE, which are commonly used in prior works:
Average Displacement Error (ADE) measures the average L2 distance between the ground truth and prediction results over all specified prediction time steps, while Final Displacement Error (FDE) represents the distance between the predicted destination and the true destination at time-step $T_{pred}$.

\noindent\textbf{Implementation Details.} 
The training was performed with Adam Optimizer with a learning rate of 0.001 and a batch size of 256. We set diffusion steps $S$ as 100. The Transformer has three layers with 512 dimensions and four attention heads. The final trajectory prediction is obtained by downsampling the Transformer output sequence through three fully connected layers (from 512d to 256d and then to 2d). %We conducted all experiments on a single NVIDIA GeForce RTX3090 GPU.

\subsection{Results and Analysis}
% \begin{figure}[!t]
% \centering
% \includegraphics[width=1\linewidth]{images/vis_eth.png}
% % \vspace{-5mm}
% \caption{The visualization result of our method on ETH/UCY datasets. Ground truths are in red solid lines, past trajectories in dark blue solid lines, and prediction results in light blue lines. We visualize 20 predictions with light blue dashed lines with each target marked with stars.}
% % \vspace{-2mm}
% \label{fig:vis}
% \end{figure}

\begin{figure*}[!t]
\centering
\includegraphics[width=1\linewidth]{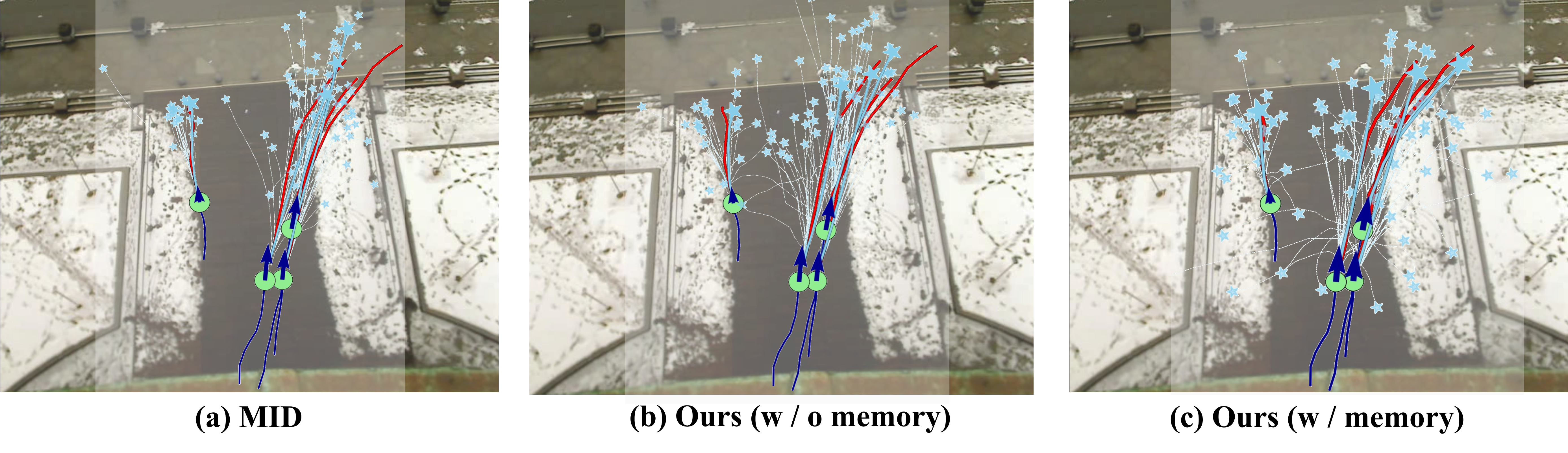}
% \vspace{-5mm}
\caption{Visualization comparison on the ETH/UCY datasets. We compare the best-of-20 predictions generated by our approach with those from two baseline methods: the previous MID method~\cite{gu2022stochastic} and our method without the motion pattern priors memory. Ground truths are in red solid lines, past trajectories in dark blue solid lines, and prediction results in light blue lines. We visualize 20 predictions for each agent with light blue dashed lines and corresponding targets are marked with stars. The result shows significant improvements by utilizing our memory-based method.}
% \vspace{-2mm}
\label{fig:vis}
\end{figure*}

\textbf{Validation Results on Pedestrian Datasests.}
We conduct a comprehensive quantitative comparison of our MP$^2$MNet method with a diverse set of contemporary approaches.
We adopt a best-of-20 evaluation strategy, consistent with previous methods~\cite{gupta2018socialgan,huang2019stgat,xu2022groupnet,gu2022stochastic} for fair comparison.

Table~\ref{sdd} provides a comparison between our method and typical existing approaches on the Stanford Drone Dataset. Our method achieves a leading average ADE of 8.89 in pixel coordinates, surpassing all prevalent approaches. Meanwhile, our method achieved the best mean ADE/FDE (0.23/0.44) for the ETH/UCY dataset, outperforming all other trajectory prediction methods. Detailed quantitative results comparing our proposed method with state-of-the-art methods are presented in Table~\ref{ethucy}.
Our method exhibits superior performance, 
demonstrating an 8\% reduction in ADE compared to the state-of-the-art GroupNet. Compared to MID~\cite{gu2022stochastic}, our method achieves improvements in both ADE and FDE in the ETH scene, with reductions of 7\% and 8\% respectively. Notably, our model exhibits significant advancements over previous methods, particularly in the ETH, UNIV, and ZARA2.
% Similarly, in the UNIV dataset, our model excels with a remarkable 19.2\% reduction in ADE and an impressive 12.2\% reduction in FDE.

\textbf{Ablation Study.} Furthermore, we explore the impact of our motion pattern priors memory bank. We substitute the target priors memory token to the raw predicted target embedding for comparison. Comparisons of the last two rows of Table~\ref{ethucy} and the visualization results in Fig.~\ref{fig:vis} (b) and (c) reveal that our memory-based approach enhances prediction performance. Our method with motion pattern priors memory outperforms the method without this memory by reducing mean ADE by 11.5\% and mean FDE by 12\%.
% Remarkably, it yields significant improvements on the ETH dataset, improving ADE by 7.1\% and FDE by 8\%.  Through the incorporation of our cluster-based memory bank into the diffusion model, we effectively retrieve potential target distributions, thereby enhancing our ability to leverage trajectory information and refine prediction results with additional target references.

\textbf{Visualization Results.}
Visualization results for comparison are tested on scenes in the ETH/UCY dataset, as shown in Fig.~\ref{fig:vis}.
% Our stochastic generative methods encourage diverse predictions. 
To validate the effectiveness of our memory approach, we compare the prediction results of our method and two other baseline methods including MID~\cite{gu2022stochastic} and our method without motion pattern priors memory. Our method has been proven to be effective based on the qualitative evaluation results. The best-of-20 prediction results are closest to the ground truth, and the predictions fall within a reasonable diverse range, which retains stochastic varieties of human movements and considers diverse human motion patterns.
% More persuasive visualization results of our method are tested on scenes in ETH/UCY dataset, as shown in Fig.~\ref{fig:vis}. Ground truths are in red solid lines, past trajectories are in dark blue solid lines, and the prediction results are in light blue lines.
% As we adopt a best-of-20 evaluation strategy following most previous methods~\cite{gupta2018socialgan,huang2019stgat,xu2022groupnet,gu2022stochastic} for a fair comparison, we visualize 20 predictions of our method with light blue dashed lines and targets for each prediction are stars. 
% Note that, stochastic generative methods are used to encourage various predictions, only the best one will be selected as the prediction result. So the best-of-20 prediction result is drawn with solid lines for metric calculation. We can find that the prediction results are reasonable and close to the ground truth within a natural uncertainty range, which validates the effectiveness of our memory, guides the diffusion model to generate trajectories following normal human motion patterns, and retains stochastic varieties.

% \vspace{-1mm}

\section{Conclusion}
\label{sec:conclusion}
In this paper, we propose %MP$^2$MNet (\textbf{M}otion \textbf{P}attern \textbf{P}riors \textbf{M}emory \textbf{Net}work), 
a novel memory approach to efficiently leverage motion pattern priors from the training set. We introduce a cluster-based memory bank to store human motion patterns with target distributions. We adopt an addressing mechanism to retrieve the matched pattern and the target distribution and generate the target priors memory token which in turn guides the diffusion model to generate trajectories. Extensive experiments and the ablation study validate the effectiveness of our method.%, achieving state-of-the-art results. % and validate the motion pattern priors memory bank, 
{\footnotesize
\bibliographystyle{IEEEbib}
\bibliography{icme2023template}}

\end{document}